%% file: root.tex
\documentclass[letterpaper, 10 pt, conference]{ieeeconf}
\IEEEoverridecommandlockouts                             
\usepackage{graphicx}
\usepackage{cite}
\usepackage[export]{adjustbox}
\usepackage{amsmath} 
\usepackage[dvipsnames]{xcolor}
\usepackage{todonotes}
\usepackage{soul}
\usepackage{balance}
\usepackage[margin=1in]{geometry} 
\usepackage[colorlinks,citecolor=blue,urlcolor=blue,bookmarks=true,hypertexnames=true]{hyperref} 
\usepackage[hypcap=true,font=footnotesize,labelfont=bf]{caption}
\usepackage{lipsum}
\usepackage{float}
\makeatletter
\def\BState{\State\hskip-\ALG@thistlm}
\makeatother 
\usepackage[dvipsnames]{xcolor}
\usepackage{subfiles}
\usepackage{subcaption}
\usepackage{algpseudocode}
\usepackage{booktabs}
\usepackage[linesnumbered,ruled,vlined]{algorithm2e} 
\usepackage{multirow} 

\title{\LARGE \bf
Exploring Kinodynamic Fabrics for Reactive Whole-Body Control of Underactuated Humanoid Robots\\[1em]
\includegraphics[width=0.7\textwidth]{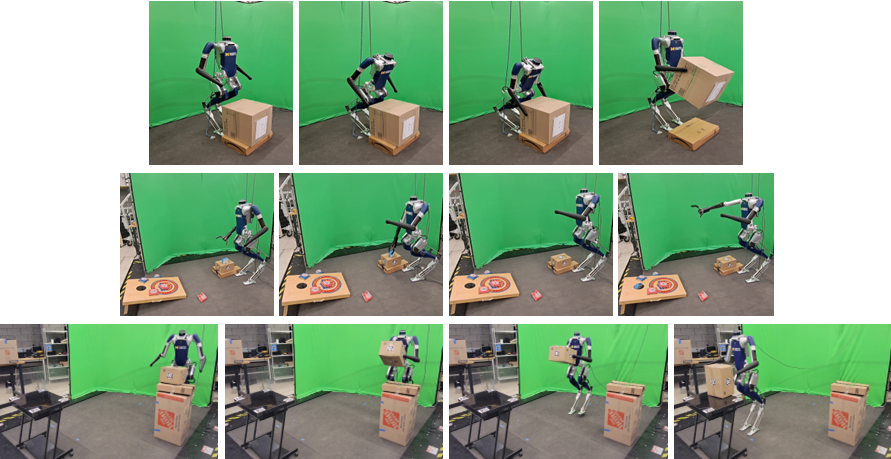} 
\captionof{figure}{A Digit robot executing a variety of whole-body motions using the Kinodynamic Fabrics Framework. (Top) Digit lifts a bulky box of non-uniform, shifting mass distribution which is not modelled in the framework. (Middle) Digit plays cornhole. (Bottom) Digit transports a package to a desired location. Digit is built by Agility Robotics.}
\label{fig:teaser}
}

\author{Alphonsus Adu-Bredu$^{*}$ \hspace{0.5cm} Grant Gibson$^{*}$ \hspace{0.5cm} Jessy Grizzle%
\thanks{All authors are with the Robotics Department, University of Michigan, Ann Arbor, MI, USA.
        {\tt\footnotesize [adubredu|grantgib|grizzle]@umich.edu}}%
\thanks{$*$ Equal Contribution}%
}

\begin{document}

\maketitle 
\setcounter{figure}{1}  
\setlength{\belowcaptionskip}{-15pt}
\thispagestyle{empty}
\pagestyle{empty}

\subfile{sections/abstract}
\subfile{sections/introduction}
\subfile{sections/related_works}

\subfile{sections/background}
\subfile{sections/problem_formulation}
\subfile{sections/methodology}

\subfile{sections/experiments}
\subfile{sections/discussion}
\subfile{sections/conclusion}


\bibliographystyle{plain}
\bibliography{references}

\balance

\end{document}

%% file: sections/abstract.tex
\begin{abstract} 
    For bipedal humanoid robots to successfully operate in the real world, they must be competent at simultaneously executing multiple motion tasks while reacting to unforeseen external disturbances in \textit{real-time}. We propose Kinodynamic Fabrics as an approach for the specification, solution and simultaneous execution of multiple motion tasks in real-time while being reactive to dynamism in the environment. Kinodynamic Fabrics allows for the specification of prioritized motion tasks as forced spectral semi-sprays and solves for desired robot joint accelerations at real-time frequencies. We evaluate the capabilities of Kinodynamic fabrics on  diverse physically-challenging whole-body control tasks with a bipedal humanoid robot both in simulation and in the real-world. Kinodynamic Fabrics outperforms the state-of-the-art Quadratic Program based whole-body controller on a variety of whole-body control tasks on run-time and reactivity metrics in our experiments. Our open-source implementation of Kinodynamic Fabrics as well as robot demonstration videos can be found at this url:
    \href{https://adubredu.github.io/kinofabs}{https://adubredu.github.io/kinofabs}
\end{abstract}

%% file: sections/introduction.tex
\section{Introduction}
The bipedal humanoid morphology is beneficial for robots for two main reasons. The first reason is that the humanoid morphology, though not specialized to traverse a particular environment, is highly agile, versatile and generalizes across very diverse environmental structures. This feature allows humans to navigate diverse environments through swimming when in water, running when on land, climbing trees and mountains and crawling under tight spaces \cite{prattvideo}.  The second reason is that if the goal is to deploy robots in human-occupied spaces, it is only natural that we fashion them to mimic human morphology to require the least amount of environmental restructuring to cater to the successful operation of robots in such spaces. Given these benefits, a bipedal humanoid robot capable of fast and reactive whole-body control would be primed to take on a diverse range of physical tasks, particularly those that may be tedious or risky for humans.

Whole-body control is the control of highly redundant, high degree-of-freedom floating-base robots to simultaneously achieve multiple motion behaviors through the exploitation of the redundancy that comes with the robot's morphology \cite{ieeeraswbc}.  The problem of fast, reactive whole-body control of bipedal humanoid robots is challenging and raises two main unanswered questions. 

The first question is, \textit{how should motion behaviors be expressed in a way that is robust and expressive}? Virtual model control has been used to create motion behaviors that rely on the design of virtual forces and components \cite{vmc} and feedback linearization has been used to track human-based motion primitives on humanoid robots for different modes of locomotion \cite{ames_mp}. A limiting factor for the use of the methods is the need for expertly designed reference motions that fit into each framework.

A popular framework for solving whole-body control problems is to express the entire controller as a constrained optimization problem that is solved in each iteration of the control loop to output joint-space commands \cite{escande2014hierarchical, salini2010lqp, del2014prioritized, herzog2014balancing, dai2014whole, feng2015optimization, koolen2016design}. In this framework, motion behaviors are expressed either  as hard constraints or as soft constraints in the objective function depending on the importance of the motion behavior. The drawback of this framework is that it mandates motion behaviors to have a specific form. For instance, if the chosen constrained optimization problem form is a Quadratic Program, motion behaviors have to strictly be linear constraints or quadratic (if appended to the objective function). It can be quite challenging to express motion behaviors such as avoiding dynamic obstacles as linear constraints without making numerous simplifying assumptions that make the problem brittle. Also, expressing multiple obstacle avoidance behaviors as constraints in a single constrained optimization problem is likely to render the problem insoluble due to potential conflicts in the constraints.


The second question is \textit{how should we formulate reactive whole-body control problems in a manner that is fast to solve and scales well with increasing number of motion behaviors}? Constrained optimization problems for whole-body control are often quite slow to solve and not amenable to real-time applications. As a result, most real-time whole-body control approaches put in the effort to make approximations to the constrained optimization problem to reduce solution time. One such approach is to formulate the constrained optimization problem as a Quadratic Program, by linearizing the constraints of the constrained optimization problem and making the objective function quadratic \cite{koolen2016design}. Besides making the problem less expressive, Quadratic Programs scale badly with increasing motion behavior constraints, resulting in slow solution times and potentially infeasible optimization problems. 

In view of these challenges, we propose Kinodynamic Fabrics as an approach for fast, reactive whole-body control of bipedal humanoid robots. Kinodynamic Fabrics allows for the description of primitive motion behaviors as forced spectral semi-sprays (fabrics) in their respective task spaces and employs the pullback and summation operations from differential geometry to compose all motion behaviors into a single joint-space motion policy. This motion policy generates joint-space acceleration commands which can be integrated into velocity and position commands and tracked by position- and velocity-controlled robots or fed as input to an inverse dynamics routine alongside desired contact forces to output joint torque commands for torque-controlled robots. The ability to express primitive motion behaviors as fabrics in their respective task spaces provides the opportunity and flexibility to express the complex geometries of smooth motion behaviors as second-order differential equations. Solving for the composed joint-space motion policy through the pullback and summation operations is very fast and allows for the computation of motion policies at kilohertz rates. These qualities make Kinodynamic Fabrics a viable framework for fast, reactive whole-body control of bipedal humanoid robots. The unique contributions of this work are as follows:
\begin{itemize}
    \item Firstly, we propose Kinodynamic Fabrics as a framework for prioritized, fast and reactive whole-body control of bipedal humanoids. We describe how to express primitive motion behaviors as expressive second-order differential equations and how they integrate into the Kinodynamic Fabrics framework to generate smooth and dynamically-consistent robot motions.
    \item Secondly, we describe how to efficiently represent complex motion behaviors like bipedal locomotion and bimanual manipulation as components of the Kinodynamic Fabrics framework, how to decompose these components into primitive motion behaviors, and how to smoothly execute extended sequences of motions.
    \item Thirdly, we demonstrate Kinodynamic Fabrics on a variety of bimanual manipulation, bipedal locomotion, and mobile manipulation tasks on the physical Digit bipedal humanoid robot.
    \item Finally, we provide an open-source Julia implementation of the Kinodynamic Fabrics framework as well as code to reproduce all of the experiments and demonstrations in this work.
\end{itemize}

We evaluate Kinodynamic Fabrics in simulations and experiments on a wide range of whole-body control tasks such as dynamic obstacle avoidance, bimanual whole-body manipulation, bipedal locomotion, and mobile manipulation, as shown in Figure \ref{fig:teaser}. In all of these tasks, Kinodynamic Fabrics generates fast, reactive motions that allow the robot to successfully accomplish the goals it is assigned. We also perform extensive benchmark comparisons of Kinodynamic Fabrics with a Quadratic Program-based whole-body controller. Kinodynamic Fabrics outperforms the Quadratic Program based controller on run-time and reactivity metrics.
\begin{figure}
    \centering
    \includegraphics[width=0.45\textwidth]{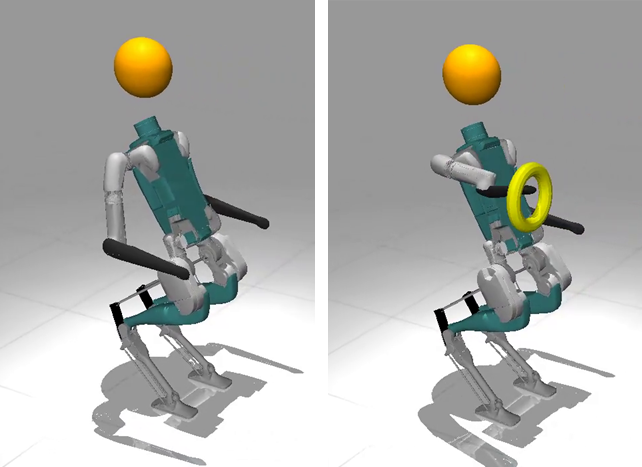}
    \caption{MuJoCo simulation of Digit executing motions generated by Kinodynamic Fabrics. (Left) Digit moves its whole body to dodge an incoming orange basketball. (Right) Digit dodges the incoming basketball while keeping the end of its right arm in the yellow hoop.}
    \label{fig:digitsim}
\end{figure}

%% file: sections/related_works.tex
\section{Related Works}
\subsection{Whole-body Control}
Whole-body control is the control of high degree-of-freedom floating-base robots to simultaneously achieve multiple motion behaviors by exploiting the redundancy that comes with the robot's structure \cite{ieeeraswbc}. Ever since whole-body control was first applied to humanoid robots in the form of Resolved Momentum Control, proposed by Kajita et. al. \cite{kajita2003resolved}, two main classes of approaches for solving whole-body control problems have been proposed; nullspace control approaches and constrained optimization-based approaches.

\subsubsection{Nullspace Whole-Body Control} 
Nullspace control  approaches \cite{kajita2003resolved, sentis2006whole, dietrich2012reactive, mistry2012operational} are dominated by the use of prioritized dynamically-consistent Jacobians and their pseudo-inverses to enforce strict hierarchies between behaviors. Lower priority behaviors have Jacobians that are defined in the nullspace of higher priority behaviors. We adopt this prioritization in Kinodynamic Fabrics for enforcing behavior priorities. The drawback of the nullspace formulation is the inability to express inequality constraints.   

\subsubsection{Constrained Optimization-based Whole-Body Control}
Constrained optimization-based whole-body control\cite{ escande2014hierarchical, salini2010lqp, del2014prioritized, feng2015optimization, dai2014whole,  koolen2016design, hopkins2016optimization} formulates the whole-body control problem as a single constrained optimization problem or cascades of constrained optimization problems. High-priority motion behaviors are written as hard constraints while low-priority behaviors are written as soft constraints in the objective function with weights to express their relative importance. Other constrained optimization-based approaches like Escande et. al. \cite{escande2014hierarchical} build a Hierarchical Quadratic Program  to enforce strict priorities between behaviors. The main drawback of constrained optimization-based whole-body control approaches is it requires non-convex formulations which require longer computation times by solvers. This computation time grows with increasing number of motion behaviors, making the approach unsuitable for real-time applications. There is, as a result, a need for linear approximations of behaviors (an example is the approximation as a Quadratic Program \cite{koolen2016design}) to speed up computations. These approximations invariably reduce the expressiveness of behaviors and increase the difficulty involved in designing motion behaviors. Another drawback of this class of approaches is that, with increasing number of motion behaviors as constraints due to, for example, increasing number of dynamic obstacles to avoid,   constrained optimization problems tend to become insoluble due to potential conflicts between the constraints.


\subsection{Optimization Fabrics}
Optimization Fabrics, or Fabrics for short, are a class of second-order differential equations called forced spectral semi-sprays \cite{ratliff2020optimization}. The second-order differential equations define smooth primitive motion behaviors and are guaranteed to optimize to a minimum when forced by a potential function. The benefit of describing primitive motion behaviors as fabrics is that we can compose different smooth behaviors from different task spaces into a single acceleration-based motion policy using pullback and summation operations from differential geometry. Geometric Fabrics, which is a class of Optimization Fabrics, has been applied to a number of reactive motion control tasks with serial manipulators \cite{xie2020geometric, van2022geometric}. Fabrics, as originally formulated, have mainly focused on the control of fully-actuated robots.  In this work, we make a number of extensions to the Optimization Fabrics framework to realize prioritized whole-body control of underactuated bipedal humanoid robots.

%% file: sections/background.tex
\section{Background}\label{sec:background}
The Kinodynamic Fabrics framework represents primitive motion behaviors as Optimization Fabrics. An Optimization Fabric, or Fabric for short, is a forced \textit{spectral semi-spray} of the form 
\begin{equation}\label{eqn:fabric}
    M \ddot{x} + f = -\delta_x \psi(x),
\end{equation}
where $M$ is a metric tensor, $f$ is a virtual force, and $-\delta_x \psi(x)$ is the applied force that pushes the system to converge to a local minimum of the potential energy function $\psi(x)$. From \eqref{eqn:fabric}, we can thus define a fabric as the second-order differential equation 
\begin{equation}
    \ddot{x} = -M^{-1} f ~-~ M^{-1}\delta_x \psi(x)
\end{equation}
and define the acceleration-based motion policy as
\begin{equation}
    \pi(x, \dot{x}) = \ddot{x}. 
\end{equation}
Each fabric component is associated with a task map, $\phi$, that maps the generalized coordinates and velocities of the robot, $q$ and $\dot{q}$, to the fabric's task space, $x$, where $x = \phi(q,\dot{q})$. Given the fabric components, the resultant joint-space acceleration is computed using pullback and summation differential geometry operations.

%% file: sections/problem_formulation.tex
\section{Problem Formulation}\label{sec:problem_formulation}
The general equations of motion of a bipedal robot can be represented as
\begin{equation}
    D(q)\ddot{q} + C(q, \dot{q})\dot{q} + G(q) = S^\top\tau + J_c^\top  F_c,
\end{equation}
 where $q, \dot{q}, \ddot{q}$ are the generalized positions, velocities, and accelerations respectively, $D(q)$ is the joint-space mass-inertia matrix, $C(q, \dot{q})$ is Coriolis matrix, $G(q)$ is the gravity vector, $S$ is the torque distribution matrix, and $\tau,\ J_c,\ F_c$ are the joint torques, end-effector contact Jacobian, and end-effector contact wrench respectively.

The Kinodynamic Fabrics framework seeks to determine the desired joint acceleration $\ddot{q}$ that simultaneously achieves a collection of motion behaviors. From $\ddot{q}$, we can then determine the torque vector $\tau$ to apply to the robot's actuators to realize the desired motion.

%% file: sections/methodology.tex
\section{Methodology}
\begin{figure}
    \centering
    \includegraphics[width=0.5\textwidth]{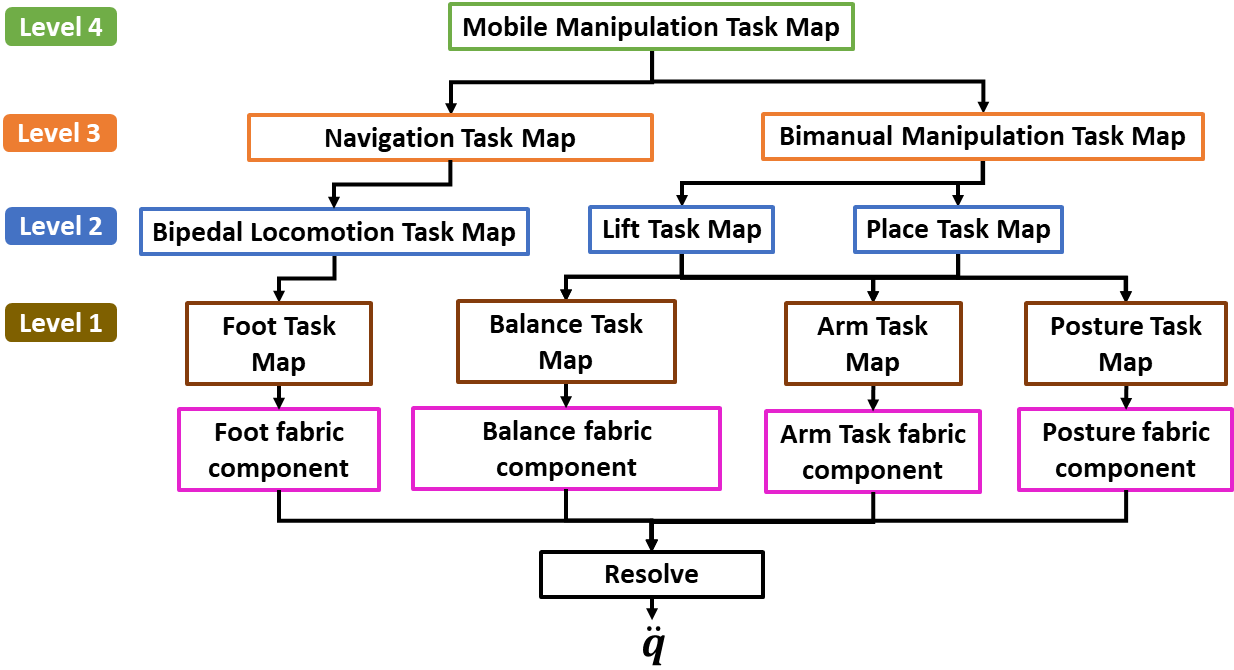}
    \caption{An illustration of the Kinodynamic Fabrics Tree for a mobile manipulation behavior. High-level behaviors compute desired target set-points for low-level behaviors. The resulting fabric components are resolved to output joint-space acceleration commands.}
    \label{fig:kinotree}
\end{figure}

Kinodynamic Fabrics takes as input the robot's generalized position and velocity vectors $q, \dot{q}$ as well as $\beta$, the collection of $K$ primitive motion behaviors whose target set-points are computed by high-level behaviors. The generalized coordinates and velocities $q, \dot{q}$ are first mapped to corresponding task space coordinates $x_k$ for each primitive motion behavior using the behavior's task map $\phi_k$. The prioritized Jacobian, $J^*_k$ of each primitive motion behavior's task map is  used to compute the derivative of the computed task space coordinates $\dot{x}_k$. Given the task space coordinates and their derivatives, $x, \dot{x}$, the task-space metric tensor $M_k(x, \dot{x})$ as well as the policy $\pi_k(x, \dot{x})$ are computed for each behavior. The pullback and summation operations from differential geometry are then applied to compose all $(M_k(x, \dot{x}), \pi_k(x, \dot{x}))$ of each motion behavior into a single joint-space acceleration vector $\ddot{q}$ \eqref{eqn:resolve}. 

The remaining parts of this section describe the various components of the framework in detail.

\subsection{Primitive Motion Behaviors}
We describe a primitive motion behavior (functionally equivalent to ``motion primitives'' in \cite{ames_mp}) as a Kinodynamic fabric represented by the tuple $(M, \pi, \phi, \rho)$ where $M$ and $\pi$ are as defined in Section \ref{sec:background} above, $\rho$ is an integer that indicates the priority level of the behavior and $\phi$ is the task map $\phi(q, \dot{q})$  that maps from the robot's configuration space position and velocity to the behavior's task space. This extension of task maps to  transform both configuration space positions and velocities to the task space is a unique feature of Kinodynamic Fabrics. 

Given $\beta$, a collection of $K$ behaviors, 
\begin{equation*}
\beta = \{(M_1, \pi_1, \phi_1, \rho_1),~ \dots, ~(M_K, \pi_K, \phi_K, \rho_K)\},
\end{equation*}
as well as the robot's generalized position and velocity vectors, $q, \dot{q}$, we solve the Kinodynamic Fabrics problem to output joint-space accelerations $\ddot{q}$. 

The theoretical convergence and stability properties of optimization fabrics have been proved in detail by Ratliff et. al. for fully-actuated systems \cite{ratliff2020optimization}. The extension of optimization fabrics for its use in the control of underactuated, hybrid systems requires a more detailed analysis, which we leave for future work. For example, the stability properties of the fabrics do not balance and walking stability guarantees of a humanoid robot. As such, the present work will mainly focus on the practical aspects of designing and implementing reactive whole-body behaviors on underactuated bipedal humanoid robots.

\subsection{Behaviors, Task Maps and Fabric Components}\label{sec:taskmaps}
Here, we describe the various classes of primitive motion behaviors we consider in this work alongside their corresponding task maps and motion policies.

The task map $\phi_k$ for a primitive motion behavior $k$ is a differentiable function that maps coordinates from the generalized coordinates and velocities $q, \dot{q}$ to the task space $x_k$. Each behavior has a unique task map that depends on the behavior's task space. 

Fabric components are motion policies that express the various behaviors described in the previous sections. Each motion policy comes along with a task-space acceleration policy $\pi(x, \dot{x})$ as well as a priority metric tensor $M(x, \dot{x})$. The priority metric tensor $M(x, \dot{x})$ is derived from a Finsler energy \cite{ratliff2020optimization, xie2020geometric}. It is an invertible matrix that encodes behavior by stretching the task space to indicate the relative priority of dimensions of the task space. The task-space acceleration policy is homogeneous of degree 2 and as such, has the form $\pi(x, \dot{x}) = -||\dot{x}||^2 \cdot \partial_x \psi(x) - B \cdot \dot{x}$ where $\psi(x)$ is a potential energy function whose local minimum satisfies task goals and $ -\partial_x \psi(x)$, the negative gradient of the potential energy function, is the force that minimizes the function and $B$ is a damping gain.

Descriptions of the classes of primitive motion behaviors for whole-body control of bipedal humanoid robots as well as their corresponding task maps, potential energy functions and priority metric tensors are as follows:

\subsubsection{Attractor Primitive Motion Behavior}\label{meth:att}
This behavior generates motions to drive a kinematic or dynamic task-space vector toward a desired value. This behavior can be used to express motions like the motion of the robot's arms, legs, body posture or center-of-mass to desired poses. 

The task map for this behavior is defined in \eqref{eq:attt}, 
where $X_g$ is the desired value and $\sigma_{att}$ is a differentiable function that maps the robot's generalized positions and velocities to the task-space vector. 
The potential energy function for this behavior is defined in \eqref{eq:attp} 
\begin{subequations}
\begin{equation}\label{eq:attt}
    \phi(q, \dot{q}) = X_g - \sigma_{att}(q, \dot{q})
\end{equation}
\begin{equation}\label{eq:attp}
    \psi(x) = \frac{1}{2}\lambda_e  x^\top x 
\end{equation}
\begin{equation}\label{eq:attm}
    M(x, \dot{x}) =W_{att}  I_n
\end{equation}
\end{subequations}
where $\lambda_e$ is a scalar gain parameter that indicates the strength of the attractive force. 

The priority metric is defined in \eqref{eq:attm}
where $W_{att}$ is a scalar weight representing the relative importance of the attractor behavior, $I_n$ is an $n \times n$ identity matrix and $n$ is the size of vector $x$.

\subsubsection{Repeller Primitive Motion Behavior}\label{meth:repeller}
This behavior generates motions to drive a kinematic or dynamic task-space vector away from an undesired value (e.g. to express obstacle or self-collision avoidance motions). For computational tractability when expressing obstacle avoidance behaviors, we specify certain finite control points on the robot's body to which this behavior is applied.

The task map for this behavior is defined in \eqref{eq:rept}  
where $X_o$ is the undesired value and  $\sigma_{rep}$ is a differentiable function that maps the robot's generalized positions and velocities to the task-space vector. 
The potential energy function for this behavior is defined in \eqref{eq:repp} 
\begin{subequations}
\begin{equation}\label{eq:rept}
    \phi(q, \dot{q}) = ||X_o - \sigma_{rep}(q, \dot{q})||^2
\end{equation}
\begin{equation}\label{eq:repp}
    \psi(x) = \frac{\lambda_b}{2} \frac{d_{\rm max} \mathbf{1} - x}{(d_{\rm max}  x) ^\top (d_{\rm max}  x)}
\end{equation}
\begin{equation}\label{eq:repm}
    M(x, \dot{x}) =  W_{rep}I_n\Big(s(\dot{x})\frac{\lambda_{om}}{x^\top x}\Big) 
\end{equation}
\end{subequations}
where $\mathbf{1}$ is a vector of ones with the same length as $x$,  $\lambda_b$ is a scalar gain parameter that indicates the strength of the repulsive force, $d_{\rm max}$ is a scalar parameter that indicates the maximum distance to the obstacle beyond which no repulsive force should be felt. 

The priority metric is defined in \eqref{eq:repm}  
where $\lambda_{om}$ is a scalar gain parameter, $W_{rep}$ is a scalar weight representing the relative importance of the repeller behavior and $s(\dot{x})$ is a velocity based switching function. $s(\dot{x}) = 1$ if $\dot{x} < 0$ and $s(\dot{x}) = 0$, otherwise. This effectively eliminates the influence of the repulsive function when the control point is moving away from the obstacle.

\subsubsection{Limit Primitive Motion Behavior}\label{meth:limit}
This behavior generates motions to keep kinematic or dynamic task-space vectors within desired limits. This is how we express inequality constraints in the Kinodynamic Fabrics framework. For example, this behavior could generate motions to keep  joint positions within their joint limits,  contact forces within friction cone limits or the zero moment point within the support polygon for robot balance regulation when standing.

The task maps for the upper and lower joint limit behaviors are defined in \eqref{eq:limt},
where $q_u$ is a vector of generalized joint upper limits and $q_l$ is a vector of generalized joint lower limits.
The potential energy function for these behaviors is defined in \eqref{eq:limp} 
\begin{subequations}
\begin{equation}\label{eq:limt}
\begin{split}
    \phi_u(q, \dot{q}) &= q_u - q \\
    \phi_l(q, \dot{q}) &= q - q_l
\end{split}
\end{equation}
\begin{equation}\label{eq:limp}
    \psi(x) = \frac{\lambda_l}{x^\top x}
\end{equation}
\begin{equation}\label{eq:limm}
    M(x, \dot{x}) =  W_{lim}I_n\Big(s(\dot{x}) \frac{\lambda_{lm}}{x^\top x}\Big)
\end{equation}
\end{subequations}
where $\lambda_l$ is a scalar gain parameter that indicates the strength of the force.

The priority metric is defined in \eqref{eq:limm}  
where $\lambda_{lm}$ is a scalar gain parameter, $W_{lim}$ is a scalar weight representing the relative importance of the limit behavior and $s(\dot{x})$ is a velocity based switching function. $s(\dot{x}) = 1$ if $\dot{x} < 0$ and $s(\dot{x}) = 0$, otherwise. This effectively eliminates the influence of the limit barrier function when the joint position is away from the joint limit.

\subsubsection{High-Level Behaviors}
Kinodynamic Fabrics also allows for the expression of high-level behaviors which determine the desired target set-points for lower-level behaviors, with primitive motion behaviors occupying the lowest level in the behavior hierarchy. We organize the behaviors into levels where primitive motion behaviors occupy level 1, higher level behaviors like locomotion and manipulation behaviors occupy level 2, and so on. The highest level of behavior is the Mobile Manipulation behavior, which takes as input a long-horizon plan made up of a sequence of high level actions (e.g. pick action, navigation action, etc) and dictates which lower-level behaviors are activated or deactivated in each iteration of the Kinodynamic Fabrics control loop. 

Only activated behaviors are evaluated in an iteration of the control loop. Once evaluated, their outputs serve as target set-points for lower-level behaviors that depend on them. The Kinodynamic Fabrics Tree represents the inter-dependence relationships between task maps of behaviors in the framework. Figure \ref{fig:kinotree} illustrates the Kinodynamic Fabrics Tree for the framework we use in our mobile manipulation experiments. Even though the task maps for primitive motion behaviors are required to be differentiable, the task maps for higher-level behaviors do not have to be differentiable.


\subsection{Prioritization}
An extension that is unique to Kinodynamic Fabrics is the use of prioritized Jacobians \cite{khatib1990motion, nemec2000null} to enforce a strict hierarchy of behaviors. Each behavior comes with a factor $\rho$ that indicates the priority of that behavior, with $\rho=1$ being the highest priority.  Lower priority behaviors have the Jacobians of their task maps defined in the nullspace of higher priority Jacobians. We denote the Jacobian $J_{\phi_{k}}$ of task map $\phi_k$ as $J_k$ for brevity. The prioritized Jacobian of behavior $k$ with priority factor $\rho$ is defined as 
\begin{equation}
    \begin{split}
        &J_k ^* = J_k \cdot S_k \cdot N_{pr(\rho)} \\
        &N_{pr(\rho)} = \prod_{j=1} ^{\rho-1} N_j \\
        &N_j = I - \Bar{J_j} J_j\\
        &\Bar{J_j} = D^{-1} J_j^T  (J_j D^{-1} J_j^T)^{-1},
    \end{split}
\end{equation}
where $pr(\rho)$ indicates behaviors that have a higher priority than $\rho$, $N_j$ is the Null-space of the behavior with priority $j$, $\Bar{J_j}$ is the dynamically-consistent pseudo-inverse of $J_j$, $D$ is the configuration space mass-inertia matrix of the robot and $S_k$ is a selection matrix that selects the actuated joints for behavior $k$.

Behaviors at the same priority are given weights $W$ on their priority metrics $M(x, \dot{x})$ to express their relative importance. In our experiments, in order to avoid discontinuities when the behavior priorities are changed only stability behaviors are priority 1. This is because stability behaviors, like the balance behavior, are invariant across different tasks and are the most critical behaviors. All other behaviors are priority 2.

\subsection{Resolution}
Having computed the task-space acceleration policy and priority metric of each behavior, we apply the pullback and summation operations to compute the desired joint-space acceleration vector,  
\begin{equation}\label{eqn:resolve}
    \ddot{q} = \left(\sum_{k=1}^K J^{*\top}_k M_k J^*_k\right)^\dagger\cdot \left(\sum_{k=1}^K J^{*\top}_k M_k(\pi_k(x_k,\dot x_k) -\dot{J}_k^*\dot{q})\right)
\end{equation}
where $(\cdot)^\dagger$ denotes the Moore-Penrose pseudo-inverse, $\ddot{q}$ is the desired actuated joint-space acceleration, and $x_k, ~\dot{x}_k$ are the task space vector and its derivative with respect to time.

%% file: sections/experiments.tex
\section{Experiments}\label{sec:experiments} 
\begin{figure*}
    \centering
    \includegraphics[width=\textwidth]{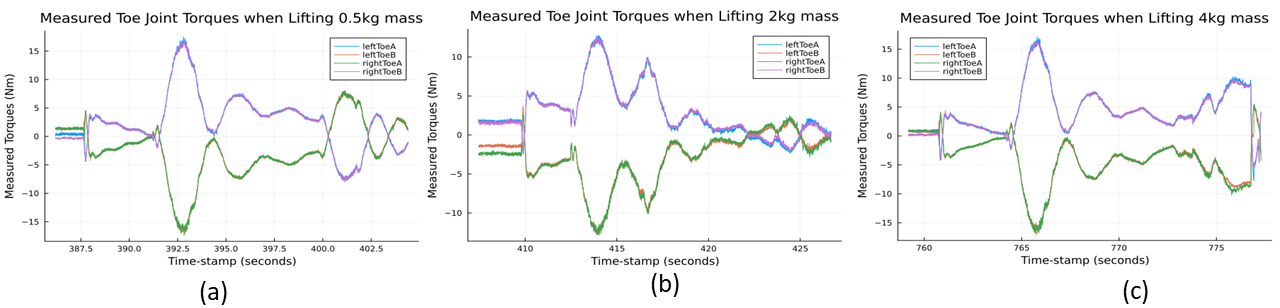}
    \caption{Plots of measured toe joint torques of the physical Digit robot while lifting large boxes. We show plots of boxes of total masses 0.5kg, 2kg and 4kg. On Digit, the bird-inspired toe acts similar to a foot on other humanoids.}
    \label{fig:bmplots}
\end{figure*}

\begin{figure}
    \centering
    \includegraphics[width=0.45\textwidth]{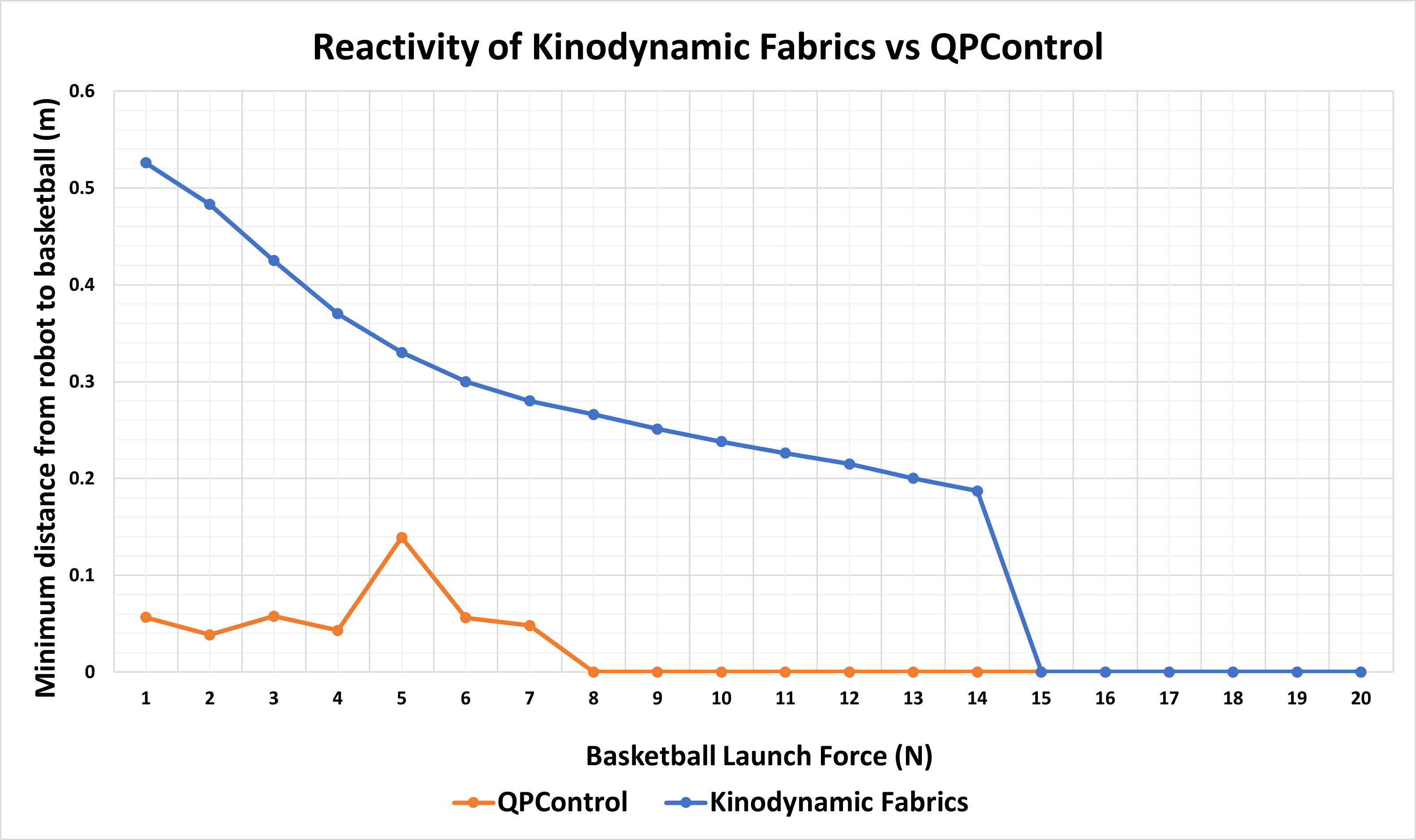}
    \caption{Plot of basketball launch forces (in Newtons) against the minimum distance of the basketball to the robot's head (in meters). A minimum distance of $0$ meters indicates that the robot fails to dodge the basketball which strikes the head of the robot and causes it to fall down. }
    \label{fig:reactivity}
\end{figure}

\begin{table}[b]
    \centering
    \resizebox{\linewidth}{!}{
    \begin{tabular}{|c||c|c|c|c|c|}
    \hline
         \textbf{Algorithm} & \textbf{PO+BL}  & \textbf{PO+EA+BL} & \textbf{PO+BL+RE} & \textbf{PO+EA+BL+RE}  \\
         \hline \hline 
         QPControl \cite{koolen2016design}& $12.60\pm2.49$  & $13.01\pm2.53$ & $13.05\pm2.48$ & $12.88\pm2.55$ \\
        \hline         
         Kinodynamic Fabrics & $0.81\pm1.18$ & $0.95\pm1.12$ & $0.92\pm1.13$ & $1.06\pm1.14$\\
         \hline
    \end{tabular}}
    \caption{Comparison of the run-time (average duration in milliseconds) of each iteration of the control loop of QPControl and Kinodynamics for different combinations of motion behaviors. PO - Whole-body Posture Behavior, EA - End-effector Attractor behavior, BL - Balance Behavior, RE - Reactivity Behavior}
    \label{table:run-time}
\end{table}

In this section, we describe various experiments we perform to evaluate the performance of Kinodynamic Fabrics on whole-body control tasks. Our experiments are performed in both simulation and in the real world on the Agility Robotics' Digit bipedal humanoid robot \cite{digit}. As shown in Figure \ref{fig:teaser}, Digit is a full bipedal humanoid robot with 30 degrees-of-freedom and 20 actuated joints. 

\subsection{Comparison with Whole-body Quadratic Program}
We compare the performance of Kinodynamic Fabrics with QPControl, a Quadratic Program whole-body control formulation proposed by Koolen et. al. \cite{koolen2016design}. We compare the performance of both approaches on whole-body control tasks using the Digit robot in a MuJoCo Physics simulation environment \cite{mujoco} as illustrated in Figure \ref{fig:digitsim}. The metrics we evaluate for both approaches are 1) \textit{Run-time}, the average duration in milliseconds of each iteration of the control loop when generating motions for different combinations of motion behaviors and 2) \textit{Reactivity}, the closest distance a dynamic obstacle, in our case, a basketball, comes to the body of the robot when the basketball is shot at the robot at different launch forces. All tasks in the following experiments have a joint-limit motion behavior to keep joint configurations within their nominal limits. The experiments were run on a Razer Blade 15 laptop with an Intel Core i7 processor with 8 cores up to 5.1GHz.

\subsubsection{Run-time Experiment}
In this experiment, we compare the duration of each iteration of the control loop for QPControl \cite{koolen2016design} and Kinodynamic Fabrics for a variety of motion behavior combinations.  The various behaviors are:
\begin{itemize}
    \item Whole-body posture behavior: This behavior is an attractor behavior (Section \ref{meth:att}) that keeps the robot in a nominal, upright posture.
    \item End-effector attractor behavior: This behavior is an attractor behavior (Section \ref{meth:att}) that keeps the left and right wrists of the robot at desired positions.
    \item Balance behavior: This behavior is a limit behavior (Section \ref{meth:limit}) that keeps the estimated zero moment point of the robot within the robot's support region. 
    \item Reactivity behavior: This behavior is a repeller behavior (Section \ref{meth:repeller}) that moves the robot's body away from an incoming basketball. 
\end{itemize}

The experimental results in Table \ref{table:run-time} indicate that Kinodynamic Fabrics is consistently faster than QPControl for all the combinations of motion behaviors.  However, the trade-off between the two approaches is that, whereas inequality constraints like joint limits or balance constraints are hard-constraints in QPControl, they can be thought of as weighted soft-constraints in Kinodynamic Fabrics. As such there may be slight violations of these constraints in Kinodynamic Fabrics.

\subsubsection{Reactivity Experiment}
This experiment compares the reactivity capabilities of QPControl and Kinodynamic Fabrics.  Specifically, the experiment evaluates how close a basketball shot at the robot at different launch forces comes to the head of the robot as the robot tries to avoid it, as depicted in Figure \ref{fig:digitsim}.  In each experiment, both approaches are tasked with a combination of the whole-body posture, end-effector attractor, balance and reactivity motion behaviors. The minimum distance of the basketball from the head of the robot is recorded. In the experiments, we vary the launch force of the basketball of mass $0.62$kg and radius $0.12$m (standard NBA ball mass and size) from $1.0$N to $20.0$N. Figure \ref{fig:reactivity} shows the experimental results for both approaches.

A general observation from the results in Figure \ref{fig:reactivity} is that, Kinodynamic Fabrics is able to effectively generate motions to keep the robot's body at a much larger distance from the basketball than QPControl. For launch forces greater than $7.0$N, QPControl, due to its slower run-time, is unable to generate motions fast enough to avoid collision with the basketball. In such situations, the basketball collides with the robot's head causing it to topple over. Kinodynamic Fabrics has a larger collision avoidance threshold of $14.0$N launch force, beyond which it collides with the basketball.

\subsection{Bimanual Manipulation and Mobile Manipulation on Physical Digit Robot}
We demonstrate the capability of Kinodynamic Fabrics to be reactive to dynamic uncertainty through bimanual manipulation tasks on the physical Digit robot, as illustrated in Figure \ref{fig:teaser}. In these tasks, Digit is made to lift bulky boxes with shifting, non-uniform mass distributions from the ground. The total masses of the boxes range from 0.5kg to 4.0kg. These masses are not modeled in the Kinodynamic Fabrics framework. From the perspective of Kinodynamic Fabrics, the shifting, non-uniform weights of the boxes are external disturbances that need to be attenuated to ensure the generation of smooth motions to keep the robot balanced while lifting the boxes.  

To keep Digit balanced while lifting the boxes, we use the balance behavior described in the previous experiment to regulate the zero moment point to stay within the support polygon. This behavior is primarily actuated by the motors in the toe joints of Digit's legs. Figure \ref{fig:bmplots} shows plots of the measured torques at the toe joints of Digit's legs as it squats and rises to lift the boxes from the ground.
As can be observed from the plots in Figure \ref{fig:bmplots}, the largest peaks in measured torques occur when Digit squats and begins to lift the box from the ground. We also demonstrate Kinodynamic Fabrics on cornhole and mobile manipulation tasks with Digit, as depicted in Figure \ref{fig:teaser}.

 For walking, we use the ALIP one-step ahead prediction method from \cite{alip} to specify components of the foot task map. Each virtual constraint is translated into an equivalent attractor primitive motion behavior. Similar set points for torso orientation and center of mass height are used and the closed-form solution for foot placement is included as a parameter in the foot task map.  Videos of all of these tasks can be found on the project webpage. 


%% file: sections/discussion.tex

%% file: sections/conclusion.tex
\section{Conclusion}
We proposed Kinodynamic Fabrics as an approach for the specification, solution, and simultaneous execution of multiple motion tasks in real-time while being reactive to dynamism in the environment. We evaluated the performance of Kinodynamic Fabrics on a variety of whole-body control tasks both in simulation and on a physical Digit robot made by Agility Robotics. Future work will integrate the Terrain-adaptive MPC formalism of \cite{alip_mpc} into the Kinodynamic Fabrics formalism. We expect it to yield more dynamic and robust locomotion of the robot, while preserving the dynamic reactivity and bimanual manipulation capabilities.

\textbf{Acknowledgements:} This work was supported in part by funds from Toyota Research Institute and the Qualcomm Innovation Fellowship. The first author thanks Prof. Nima Fazeli for useful discussions on Geometric Fabrics. All of the authors thank Prof. Odest Chadwicke Jenkins for his encouragement and insight on mobile manipulation.

%% file: root.bbl
\begin{thebibliography}{10}

\bibitem{dai2014whole}
Hongkai Dai, Andr{\'e}s Valenzuela, and Russ Tedrake.
\newblock Whole-body motion planning with centroidal dynamics and full
  kinematics.
\newblock In {\em 2014 IEEE-RAS International Conference on Humanoid Robots},
  pages 295--302. IEEE, 2014.

\bibitem{del2014prioritized}
Andrea Del~Prete, Francesco Romano, Lorenzo Natale, Giorgio Metta, Giulio
  Sandini, and Francesco Nori.
\newblock Prioritized optimal control.
\newblock In {\em 2014 IEEE International Conference on Robotics and Automation
  (ICRA)}, pages 2540--2545. IEEE, 2014.

\bibitem{ieeeraswbc}
Alexander Dietrich, Jaeheung Park, Luis Sentis, and Niels Dehio.
\newblock Technical committee for whole-body control.
\newblock {\em IEEE Robotics and Automation Society}, 2014.

\bibitem{dietrich2012reactive}
Alexander Dietrich, Thomas Wimbock, Alin Albu-Schaffer, and Gerd Hirzinger.
\newblock Reactive whole-body control: Dynamic mobile manipulation using a
  large number of actuated degrees of freedom.
\newblock {\em IEEE Robotics \& Automation Magazine}, 19(2):20--33, 2012.

\bibitem{escande2014hierarchical}
Adrien Escande, Nicolas Mansard, and Pierre-Brice Wieber.
\newblock Hierarchical quadratic programming: Fast online humanoid-robot motion
  generation.
\newblock {\em The International Journal of Robotics Research},
  33(7):1006--1028, 2014.

\bibitem{feng2015optimization}
Siyuan Feng, Eric Whitman, X~Xinjilefu, and Christopher~G Atkeson.
\newblock Optimization-based full body control for the darpa robotics
  challenge.
\newblock {\em Journal of field robotics}, 32(2):293--312, 2015.

\bibitem{alip_mpc}
Grant Gibson, Oluwami Dosunmu-Ogunbi, Yukai Gong, and Jessy Grizzle.
\newblock Terrain-adaptive, alip-based bipedal locomotion controller via model
  predictive control and virtual constraints.
\newblock In {\em 2022 IEEE/RSJ International Conference on Intelligent Robots
  and Systems (IROS)}, pages 6724--6731, 2022.

\bibitem{alip}
Yukai Gong and Jessy~W. Grizzle.
\newblock Zero dynamics, pendulum models, and angular momentum in feedback
  control of bipedal locomotion.
\newblock {\em Journal of Dynamic Systems, Measurement, and Control}, 144(12),
  10 2022.
\newblock 121006.

\bibitem{herzog2014balancing}
Alexander Herzog, Ludovic Righetti, Felix Grimminger, Peter Pastor, and Stefan
  Schaal.
\newblock Balancing experiments on a torque-controlled humanoid with
  hierarchical inverse dynamics.
\newblock In {\em 2014 IEEE/RSJ International Conference on Intelligent Robots
  and Systems}, pages 981--988. IEEE, 2014.

\bibitem{hopkins2016optimization}
Michael~A Hopkins, Alexander Leonessa, Brian~Y Lattimer, and Dennis~W Hong.
\newblock Optimization-based whole-body control of a series elastic humanoid
  robot.
\newblock {\em International Journal of Humanoid Robotics}, 13(01):1550034,
  2016.

\bibitem{kajita2003resolved}
Shuuji Kajita, Fumio Kanehiro, Kenji Kaneko, Kiyoshi Fujiwara, Kensuke Harada,
  Kazuhito Yokoi, and Hirohisa Hirukawa.
\newblock Resolved momentum control: Humanoid motion planning based on the
  linear and angular momentum.
\newblock In {\em Proceedings 2003 IEEE/RSJ International Conference on
  Intelligent Robots and Systems (IROS 2003)(cat. no. 03ch37453)}, volume~2,
  pages 1644--1650. IEEE, 2003.

\bibitem{khatib1990motion}
Oussama Khatib.
\newblock Motion/force redundancy of manipulators.
\newblock In {\em Proceedings of Japan-USA Symposium on Flexible Automation},
  volume~1, pages 337--342, 1990.

\bibitem{koolen2016design}
Twan Koolen, Sylvain Bertrand, Gray Thomas, Tomas De~Boer, Tingfan Wu, Jesper
  Smith, Johannes Englsberger, and Jerry Pratt.
\newblock Design of a momentum-based control framework and application to the
  humanoid robot atlas.
\newblock {\em International Journal of Humanoid Robotics}, 13(01):1650007,
  2016.

\bibitem{mistry2012operational}
Michael Mistry and Ludovic Righetti.
\newblock Operational space control of constrained and underactuated systems.
\newblock In {\em Robotics: Science and systems}, volume~7, pages 225--232,
  2012.

\bibitem{nemec2000null}
Bojan Nemec and Leon Zlajpah.
\newblock Null space velocity control with dynamically consistent
  pseudo-inverse.
\newblock {\em Robotica}, 18(5):513--518, 2000.

\bibitem{ames_mp}
Matthew~J. Powell, Huihua Zhao, and Aaron~D. Ames.
\newblock Motion primitives for human-inspired bipedal robotic locomotion:
  walking and stair climbing.
\newblock In {\em 2012 IEEE International Conference on Robotics and
  Automation}, pages 543--549, 2012.

\bibitem{vmc}
J.~Pratt, P.~Dilworth, and G.~Pratt.
\newblock Virtual model control of a bipedal walking robot.
\newblock In {\em Proceedings of International Conference on Robotics and
  Automation}, volume~1, pages 193--198 vol.1, 1997.

\bibitem{prattvideo}
Jerry Pratt.
\newblock Toward humanoid avatar robots for co-exploration of hazardous
  environments.
\newblock \url{https://youtu.be/HefjKANiZx0}, 2018.

\bibitem{ratliff2020optimization}
Nathan~D Ratliff, Karl Van~Wyk, Mandy Xie, Anqi Li, and Muhammad~Asif Rana.
\newblock Optimization fabrics.
\newblock {\em arXiv preprint arXiv:2008.02399}, 2020.

\bibitem{digit}
Agility Robotics.
\newblock Digit robot.
\newblock \url{https://www.agilityrobotics.com/meet-digit}.

\bibitem{salini2010lqp}
Joseph Salini, S{\'e}bastien Barth{\'e}lemy, and Philippe Bidaud.
\newblock Lqp controller design for generic whole body motion.
\newblock In {\em Mobile Robotics: Solutions and Challenges}, pages 1081--1090.
  World Scientific, 2010.

\bibitem{sentis2006whole}
Luis Sentis and Oussama Khatib.
\newblock A whole-body control framework for humanoids operating in human
  environments.
\newblock In {\em Proceedings 2006 IEEE International Conference on Robotics
  and Automation, 2006. ICRA 2006.}, pages 2641--2648. IEEE, 2006.

\bibitem{mujoco}
Emanuel Todorov, Tom Erez, and Yuval Tassa.
\newblock Mujoco: A physics engine for model-based control.
\newblock In {\em 2012 IEEE/RSJ International Conference on Intelligent Robots
  and Systems}, pages 5026--5033. IEEE, 2012.

\bibitem{van2022geometric}
Karl Van~Wyk, Mandy Xie, Anqi Li, Muhammad~Asif Rana, Buck Babich, Bryan Peele,
  Qian Wan, Iretiayo Akinola, Balakumar Sundaralingam, Dieter Fox, et~al.
\newblock Geometric fabrics: Generalizing classical mechanics to capture the
  physics of behavior.
\newblock {\em IEEE Robotics and Automation Letters}, 7(2):3202--3209, 2022.

\bibitem{xie2020geometric}
Mandy Xie, Karl Van~Wyk, Anqi Li, Muhammad~Asif Rana, Qian Wan, Dieter Fox,
  Byron Boots, and Nathan Ratliff.
\newblock Geometric fabrics for the acceleration-based design of robotic
  motion.
\newblock {\em arXiv preprint arXiv:2010.14750}, 2020.

\end{thebibliography}
